\documentclass[journal]{IEEEtran}
\usepackage{graphicx}
\usepackage{hyperref}
\usepackage{amsmath,amssymb,amsfonts}
\usepackage{algorithmic}
\usepackage{algorithm}
\usepackage{multirow}
\usepackage{booktabs}

\usepackage{orcidlink}
\usepackage{xcolor}

\graphicspath{{figures/}}

\DeclareMathOperator*{\argmin}{arg\,min}

\begin{document}
\title{S3D-Mosaic: Subgradient-Guided Cross-Band Structure Transfer for Spectral Demosaicing}

\author{Weiqiang~Zhao\orcidlink{0009-0009-2714-4622}, 
        Xian~Li\orcidlink{0000-0001-5714-3940},~\IEEEmembership{Member,~IEEE,}
        Yuzhe~Gui\orcidlink{0009-0007-2630-6736},
        Tianzhu~Liu\orcidlink{0000-0001-6903-9614},~\IEEEmembership{Member,~IEEE,}
        Yanfeng~Gu\orcidlink{0000-0003-1625-7989},~\IEEEmembership{Senior~Member,~IEEE}
\thanks{This work was supported in part by the National Natural Science Foundation of China through the Major Scientific Instrument
Development Program under Grant 62327803. (Corresponding author: Yanfeng Gu).}
\thanks{Weiqiang Zhao, Xian Li, Yuzhe Gui, Tianzhu Liu and Yanfeng Gu are with the School of Electronics and Information Engineering, Harbin Institute of Technology, Harbin 150001, China (E-mail: wqzhao@stu.hit.edu.cn; xianli@hit.edu.cn; 25B305002@stu.hit.edu.cn; tzliu@hit.edu.cn;guyf@hit.edu.cn).}}

\markboth{Journal of \LaTeX\ Class Files,~Vol.~14, No.~8, August~2021}%
{Zhao \MakeLowercase{\textit{et al.}}: S3D-Mosaic Spectral Demosaicing}


\maketitle

\begin{abstract}
Multispectral filter array (MSFA) imaging enables snapshot spectral acquisition, but its per-band sparse sampling causes severe spatial aliasing. 
Although missing spatial structures can be inferred across bands, wavelength-dependent contrast variations make direct cross-band transfer unreliable. 
Existing model-based methods lack explicit magnitude-invariant mechanisms, and learning-based methods require layout-specific training or per-image tuning, limiting generalization. 
To address these issues, we propose S3D-Mosaic, a model-based spectral demosaicing method built upon a subgradient-guided cross-band structure transfer framework.
We first establish a subgradient-guided regularization framework with a generalized cosine-similarity interpretation, providing a convex mechanism for magnitude-invariant information transfer.
Then we design a compact Directional Sparse Equivariant Difference Convolution (DSE-DConv) for band-level extraction of orientation-selective, high-frequency structure features at low computational cost.
Instantiating the framework with DSE-DConv on spectrally adjacent bands yields a cross-band self-structure similarity regularizer for spectral demosaicing, which we optimize via a dual-scan ADMM algorithm that decouples the bidirectional dependency while propagating magnitude-robust structural information across bands.
With DSE-DConv's two parameters trained once via self-supervision, S3D-Mosaic outperforms state-of-the-art unsupervised methods on CAVE and NTIRE 2022, and generalizes robustly to varying MSFA layouts, pan-mosaic fusion, and real-world UAV data with fixed reconstruction hyperparameters.
The source code is available at \url{https://github.com/bestwishes43/S3D-Mosaic}.
\end{abstract}


\section{Introduction}
\IEEEPARstart{S}{napshot} spectral imaging provides an effective way to capture three-dimensional spatio-spectral information in a single exposure. Among various hardware realizations, multispectral filter array (MSFA) cameras overlay a spatially repeating spectral filter on a conventional sensor, achieving a compact, cost-effective form factor suited to applications including material identification~\cite{lin_underdetermined_2026}, biophysical parameter retrieval~\cite{cao_improved_2025}, and 3D reconstruction~\cite{wang_robust_2023, wang_adaptive_2024}.

However, this design trades spatial resolution for spectral diversity: each band is observed only at a sparse set of positions, causing high-frequency spatial components to alias into lower frequencies.
These artifacts are hard to remove in scenes with sharp boundaries or fine textures, because the lost high-frequency information cannot be uniquely recovered from a single band.
A key observation is that spatial discontinuities, such as object boundaries, shadows, and surface geometry, are shared across most spectral bands, and the MSFA's sampling means details lost in one band are often preserved in others. This suggests that missing high-frequency structures can be inferred across bands. Yet cross-band transfer is not straightforward: corresponding edges often exhibit markedly different contrast across wavelengths, so directly matching intensities or gradients may distort spectral information.

Existing model-based methods exploit global spectral correlations via sparsity~\cite{elad_image_2006} or low-rankness~\cite{candes_exact_2009,qin_lowrank_2022}, or encourage cross-band similarity by minimizing intensity/gradient differences~\cite{mizutani_multispectral_2014}. While interpretable, they may transfer spectral contrast rather than spatial structure, and their priors often provide insufficient constraints for heavily aliased high-frequency content.
Learning-based methods, including supervised networks~\cite{feng_mosaic_2021,wu_mosaic_2025} and unsupervised approaches~\cite{feng_unsupervised_2024,chen_unsupervised_2025}, learn expressive models from data. However, their performance can depend on the training data distribution, the MSFA layout used during training, or per-image hyperparameter tuning~\cite{chen_unsupervised_2025}. This is especially problematic for spectral demosaicing, given the diversity of MSFA layouts and the cost of paired training data.

\begin{figure*}
\centering
\includegraphics[width=0.90\textwidth]{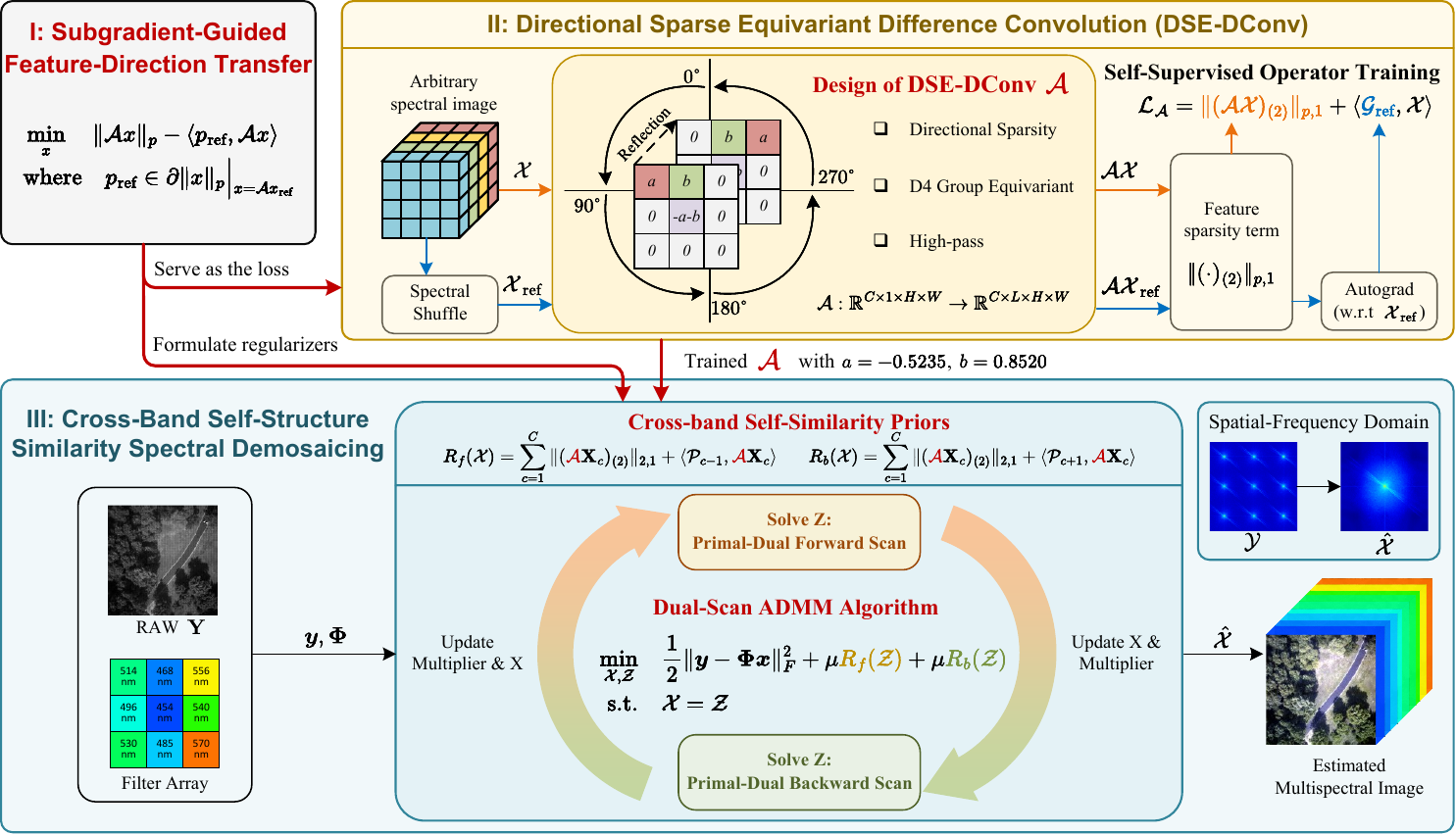}
\caption{Illustration of the proposed S3D-Mosaic pipeline: (i) a subgradient-guided optimization framework is developed to realize magnitude-invariant cross-band feature-direction transfer; (ii) the compact DSE-DConv operator is designed and self-supervisedly trained as a structural feature extractor; (iii) a cross-band self-structure similarity prior is instantiated for spectral demosaicing and optimized using a dual-scan ADMM algorithm.}
\label{fig:illustration}
\end{figure*}

To this end, we propose S3D-Mosaic, a model-based method that realizes magnitude-robust cross-band structure transfer through subgradient guidance.
Instead of matching magnitudes, it aligns feature directions through the subgradient of a norm-based regularizer, giving a convex, magnitude-invariant mechanism with a generalized cosine-similarity interpretation.
To instantiate the framework, we further design Directional Sparse Equivariant Difference Convolution (DSE-DConv), a compact two-parameter linear operator that extracts orientation-selective high-frequency structural features with $D_4$ transformation consistency. 
Using DSE-DConv, we construct a cross-band self-structure similarity prior between adjacent spectral bands and solve the resulting problem through a dual-scan ADMM algorithm.
Fig.~\ref{fig:illustration} overviews the pipeline.

The main contributions are summarized as follows:

\begin{enumerate}
\item We establish a subgradient-guided feature-direction transfer framework with a generalized cosine-similarity interpretation, which provides a principled mechanism for magnitude-invariant information transfer.
\item We design DSE-DConv, a two-parameter \(D_4\)-equivariant zero-sum operator, to extract transformation-consistent directional high-frequency structures for demosaicing.
\item We instantiate the framework with DSE-DConv as a cross-band self-structure similarity prior and develop a dual-scan ADMM solver to decouple bidirectional spectral dependencies. The resulting method generalizes to diverse MSFA layouts and real captures with fixed hyperparameters.
\end{enumerate}

\IEEEpubidadjcol
The remainder of this paper is organized as follows. 
Section~\ref{sec:related} reviews related work; Section~\ref{sec:model} formulates the problem; Sections~\ref{sec:theory}--\ref{sec:algorithm} present the proposed framework, DSE-DConv operator, and optimization algorithm; Section~\ref{sec:experiments} reports experiments; and Section~\ref{sec:conclusion} concludes the paper.

\section{Related Work}
\label{sec:related}
The proposed S3D-Mosaic is related to spectral demosaicing, cross-band self-similarity priors, and low-level structural feature extraction. We review these three aspects below.
\subsection{Spectral demosaicing}
Existing methods can be broadly divided into model-based and learning-based approaches.
model-based methods solve the reconstruction problem with explicit image priors. 
Early methods relied on spatial coherence in a single band, including bilinear interpolation~\cite{brauers_color_2006}, edge-sensing interpolation~\cite{miao_binary_2006}, and diffraction-limited interpolation~\cite{yasuma_generalized_2010}. Although efficient, local interpolation is often insufficient for removing structured aliasing caused by severe high-frequency undersampling. 
Later methods, such as SD~\cite{brauers_color_2006} and ItSD~\cite{mizutani_multispectral_2014}, exploit spectral correlation among bands, while PPID~\cite{mihoubi_multispectral_2015,mihoubi_multispectral_2017} constructs a pseudo-panchromatic image to provide stronger structural guidance. 
Later studies formulated spectral demosaicing as a regularized inverse problem using dictionary sparsity~\cite{degraux_generalized_2015}, spectral low-rankness~\cite{tsagkatakis_graph_2019,bian_generalized_2021}, spatial smoothness, or hybrid priors~\cite{zhuang_hydemosaicing_2022,abbas_locallyrankonebased_2024}. 
Recently, Li \textit{et al.}~\cite{li_efficient_2025} proposed a convex formulation that combines spectral low-rankness and spatial smoothness and shows robustness to different MSFA layouts. 
Although interpretable and requiring little or no training data, they do not explicitly address the key difficulty of transferring aliasing-corrupted structures across bands when corresponding edges have wavelength-dependent magnitudes or contrasts.

Learning-based methods learn either the reconstruction mapping or the image prior from data. 
Supervised networks, including attention-based modules~\cite{feng_mosaic_2021}, spatial-frequency representation networks~\cite{wu_mosaic_2025,wu_joint_2025,xu_waveformer_2025}, and multi-stage restoration networks~\cite{xiong_multistage_2025}, rely on paired data for training.
Unsupervised methods reduce the data dependence; USD~\cite{feng_unsupervised_2024} leverages an equivariant imaging prior for unsupervised reconstruction network training, while UnNull~\cite{chen_unsupervised_2025} performs reconstruction via Deep Image Prior (DIP)~\cite{lempitsky_deep_2018}.
Despite their strong performance, these methods may be sensitive to training distributions, MSFA layouts, or per-scene optimization settings.

In contrast, S3D-Mosaic maintains an explicit reconstruction model and uses learning only to calibrate a compact structural feature operator.
\subsection{Cross-Band Self-Similarity Priors}
Self-similarity is a fundamental image prior. 
Classical non-local methods, such as NL-means~\cite{buades_nonlocal_2005} and BM3D~\cite{dabov_image_2007}, exploit repeated patches within an image, while BM4D~\cite{maggioni_nonlocal_2013} and non-local low-rank models~\cite{chang_hyperlaplacian_2017,he_nonlocal_2022} extend this idea to volumetric or spectral data. 
However, these methods usually measure similarity at the intensity or patch level and do not explicitly separate structural direction from feature magnitude.

For spectral images, cross-band consistency provides another important form of self-similarity. 
Spectral-spatial TV variants penalize local spatial-spectral variations~\cite{yuan_hyperspectral_2012,chen_hyperspectral_2023,zhao_spatialspectral_2026}, while low-rank priors encourage global spectral coherence~\cite{tsagkatakis_graph_2019,bian_generalized_2021,he_nonlocal_2022}. 
Explicit structural guidance can also be constructed from the spectral data itself, such as pseudo-panchromatic images~\cite{mihoubi_multispectral_2015,mihoubi_multispectral_2017} or estimated green channels~\cite{guo_joint_2021,kong_image_2025}. 
These strategies demonstrate the value of cross-band guidance, but the guidance is commonly imposed through intensity or gradient consistency, making it sensitive to wavelength-dependent contrast variations.

Recent deep models also inject self-similarity implicitly or explicitly. Non-local networks~\cite{wang_nonlocal_2018,liu_nonlocal_2018}, Vision Transformers~\cite{dosovitskiy_image_2021}, and state-space models~\cite{liu_vmamba_2024} can learn long-range or cross-channel dependencies from data, while Wu \textit{et al.}~\cite{wu_mosaic_2025} exploit self-similarity through spatial-spectral attention and PPI guidance, and UnNull~\cite{chen_unsupervised_2025} combines an online DIP with spectral-smoothness regularization. However, these deep-learning methods typically learn the cross-band relationship implicitly from training data, without enforcing an explicit structural alignment mechanism.

In contrast, S3D-Mosaic formulates cross-band self-similarity using a convex subgradient-guided transfer mechanism: it transfers the feature direction via a dual subgradient, providing a principled way to propagate structural information robust to intensity variations across bands.

\subsection{Low-Level Structure Extraction}
Classical filters such as Gabor, log-Gabor, and steerable filters explicitly encode oriented, localized, and band-pass responses. 
The relevance of such responses to efficient visual representation is supported by sparse-coding studies of natural images~\cite{olshausen_emergence_1996}, but these filters often require large filter banks and careful parameter selection.

Learnable convolutions have been developed to incorporate useful inductive biases into neural operators. Difference-based convolutions, including Local Binary Patterns~\cite{liu_extended_2012,liu_median_2016}, Central Difference Convolution~\cite{yu_searching_2020,yu_nasfas_2021}, and Pixel Difference Convolution~\cite{su_lightweight_2023}, enhance high-frequency and edge-sensitive responses by introducing local intensity differences. Group-equivariant convolutions enforce consistency under transformations such as rotation and reflection~\cite{cohen_group_2016}, and later works extend equivariance to continuous or affine transformations~\cite{xie_fourier_2023,li_affine_2024}. Directional sparse representations, such as steerable pyramids~\cite{freeman_design_1991} and strip-like directional convolutions~\cite{xu_infrared_2023}, aim to isolate orientation-specific responses and reduce cross-orientation interference.
These works provide high-pass sensitivity, transformation equivariance, or directional selectivity from different perspectives, but few are tailored to model-based cross-band structure transfer. 
DSE-DConv combines these properties in a compact linear operator with only two learnable parameters.

\section{Problem Formulation}
\label{sec:model}

Following the standard MSFA observation model \cite{brauers_color_2006, yasuma_generalized_2010, degraux_generalized_2015}, the sensor records a single-channel mosaic image \(\mathbf{Y}\in\mathbb{R}^{H\times W}\) in which each pixel samples one of \(C\) spectral bands according to a predefined spatial layout \(\mathbf{T}\in\{1,\dots,C\}^{H\times W}\). Let \(\mathcal{X}\in\mathbb{R}^{C\times H\times W}\) denote the fully-sampled multispectral image. 
The observation can be expressed via a binary mask tensor \(\mathcal{M}\in\{0,1\}^{C\times H\times W}\), where \(\mathcal{M}_{c,h,w}=1\) if \(\mathbf{T}_{h,w}=c\) and zero otherwise:
\begin{equation}
\mathbf{Y} = \sum_{c=1}^{C} \mathbf{M}_{c} \odot \mathbf{X}_{c},
\label{eq:mosaic_sum}
\end{equation}
where \(\odot\) denotes element-wise multiplication and \(\mathbf{X}_{c}\in \mathbb{R}^{H \times W}\), \(\mathbf{M}_{c}\in \mathbb{R}^{H \times W}\) are the \(c\)-th band slices of \(\mathcal{X}\) and \(\mathcal{M}\), respectively.
Vectorizing Eq.~\eqref{eq:mosaic_sum}, we obtain the linear model
\begin{equation}
\boldsymbol{y} = \sum_{c=1}^{C} \mathbf{\Phi}_c \boldsymbol{x}_c = \mathbf{\Phi} \boldsymbol{x},
\label{eq:vec_obs}
\end{equation}
where \(\boldsymbol{y}, \boldsymbol{x}_c\) are the vectorized versions of \(\mathbf{Y}\) and \(\mathbf{X}_{c}\), respectively; \(\mathbf{\Phi}_c=\mathrm{diag}(\mathrm{vec}(\mathbf{M}_c))\); and \(\mathbf{\Phi} = [\mathbf{\Phi}_1 \, \cdots \, \mathbf{\Phi}_C]\), \(\boldsymbol{x} = [\boldsymbol{x}_1^\top \, \cdots \, \boldsymbol{x}_C^\top]^\top\).

It is easy to check that the mask matrices satisfy \(\mathbf{\Phi}_i\mathbf{\Phi}_j = \mathbf{\Phi}_i\) if \(i=j\) and zero otherwise, hence \(\mathbf{\Phi}\mathbf{\Phi}^\top = \mathbf{I}\). 
This orthogonal-diagonal structure implies that multiplying \eqref{eq:vec_obs} on the left by \(\mathbf{\Phi}_c^\top\) simply extracts the observed pixels of band \(c\) without loss of information. In tensor form, this equivalent observation \(\mathcal{Y}\in \mathbb{R}^{C\times H \times W}\) reads
\[
\mathcal{Y}_{c,:,:} = \mathbf{M}_c \odot \mathbf{Y} = \mathbf{M}_c \odot \mathbf{X}_c.
\]

\begin{figure}[ht]
    \centering
    \includegraphics[width=\linewidth]{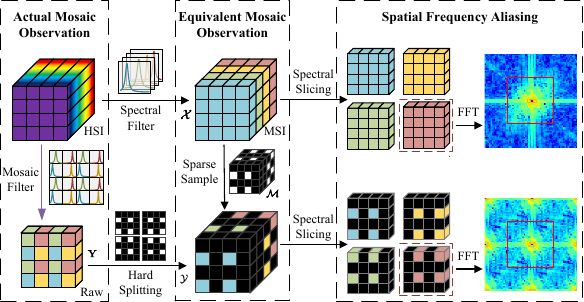}
    \caption{Illustration of the MSFA observation. Left: actual mosaic observation model from hyperspectral image (HSI) to raw; Middle: equivalent per-band subsampling model; Right: spatial aliasing caused by sparse sampling.}
    \label{fig:obs_model}
\end{figure}

It reveals that each spectral band \(\mathbf{X}_c\) is sampled only at the spatial positions where \(\mathcal{M}_{c,h,w} = 1\).
The subsampling leads to severe spatial frequency aliasing, as illustrated in Fig.~\ref{fig:obs_model}, and makes the recovery of \(\mathcal{X}\) highly ill-posed. 

When the scene contains rich high-frequency content, conventional interpolation cannot eliminate aliasing or recover the lost high-frequency information; additional prior knowledge must therefore be exploited. 
Assuming additive white Gaussian noise (e.g., photon shot noise, readout noise), the spectral demosaicing problem can be formulated as:
\begin{equation}
\min_{\mathcal{X}} \; \frac{1}{2} \left\|\boldsymbol{y} - \mathbf{\Phi} \boldsymbol{x}\right\|_2^2 + \mu R(\mathcal{X}),
\label{eq:inv_prob}
\end{equation}
where the regularizer \(R(\mathcal{X})\) encodes prior knowledge about \(\mathcal{X}\); the \(\ell_2\)-norm based data fidelity term corresponds to a Gaussian noise assumption.

\section{Subgradient-Guided Feature-Direction Transfer}
\label{sec:theory}
Cross-band structural cues can compensate for the undersampling shown in Section~\ref{sec:model}, but directly matching intensities or gradients risks transferring spectral contrast. To avoid this, we use the subgradient of a norm-based regularizer to encode feature directions, providing a convex, magnitude-invariant information transfer mechanism.

\subsection{Subgradient-Guided Regularization with Linear Operators}
We first present a general construction that explains how information can be transferred through subgradients.

Let \(f: \mathbb{E} \to \bar{\mathbb{R}}\) be a proper, closed, convex function and let \(g_y \in \partial f(y)\) be a subgradient at a reference point \(y \in \operatorname{dom} f\).  
Consider the optimization problem
\begin{equation}
\min_{x} \; \bigl\{ f(x) - \langle g_y, x \rangle \bigr\}.
\label{eq:subgrad_opt}
\end{equation}

By convex analysis, any minimizer \(x^*\) of \(\min_x\{f(x)-\langle g_y,x\rangle\}\) satisfies \(g_y\in\partial f(x^*)\) and thus shares the subgradient with \(y\)~\cite{zhao_tv_2026}. In contrast to the specific function setting in \cite{zhao_tv_2026}, we adopt a more general composite form \(f = h \circ \mathcal{A}\), where \(\mathcal{A}\) is a bounded linear operator that extracts features, and \(h: \mathbb{E}_2 \to \bar{\mathbb{R}}\) is a closed convex function that measures deviation from an ideal feature set.  
The subdifferential chain rule~\cite{Rockafellar1970} gives
\[
\partial f(x) = \mathcal{A}^* \partial h(\mathcal{A}x),
\]
where \(\mathcal{A}^*\) is the adjoint of \(\mathcal{A}\).
Hence, for any \(g_y \in \partial f(y)\) there exists a dual subgradient \(p_y \in \partial h(\mathcal{A}y)\) such that \(g_y = \mathcal{A}^* p_y\).  
Define the subgradient-guided functional
\begin{equation}
D_{f}(x; y) \triangleq f(x) - \langle \mathcal{A}^* p_y, x \rangle 
= h(\mathcal{A}x) - \langle p_y, \mathcal{A}x \rangle .
\label{eq:D_f}
\end{equation}

By the optimality condition, any minimizer \(x^*\) of Eq.~\eqref{eq:D_f} satisfies \(p_y \in \partial h(\mathcal{A}x^*)\), so the information encoded in the dual subgradient \(p_y\) constrains the feature representation \(\mathcal{A}x^*\).

\subsection{Magnitude-Invariant Directional Alignment}
To make the transfer independent of the reference feature's magnitude, we let \(h\) be a 1-homogeneous function.
A canonical choice is the \(\ell_p\) norm (\(p \ge 1\)), which yields \(f(x) = \|\mathcal{A}x\|_p\). Notably, with an appropriate linear operator \(\mathcal{A}\), this formulation can express sparsity, smoothness, or low-rank regularization.

For \(v \in \mathbb{E}_2\), the subdifferential of \(h\) is
\begin{equation}
\partial h(v) =
\begin{cases}
\left\{\dfrac{\psi_p(v)}{\|\psi_p(v)\|_q}\right\}, & v \neq 0,\\[1.2em]
\{\, u : \|u\|_q \le 1 \,\}, & v = 0,
\end{cases}
\label{eq:subdiff_p}
\end{equation}
where \(q\) is the Hölder conjugate of \(p\) (\(1/p+1/q=1\)) and \(\psi_p\) is the duality mapping
\[
(\psi_p(v))_i = \operatorname{sgn}(v_i)\,|v_i|^{p-1}.
\]

Substituting \(h\) into~\eqref{eq:D_f} gives
\begin{equation}
D_{f}(x; y) = \|\mathcal{A}x\|_p - \langle p_y, \mathcal{A}x \rangle=\langle p_x-p_y, \mathcal{A}x \rangle.
\label{eq:Df_diff}
\end{equation}

This structure motivates the following definition of the \(\ell_p\)-norm-induced generalized cosine similarity:
\begin{equation}
\mathrm{CosSim}_p(u; v) \triangleq \left\langle \frac{u}{\|u\|_p},\,
\frac{\psi_p(v)}{\|\psi_p(v)\|_q} \right\rangle,
\label{eq:Cp}
\end{equation}
which satisfies \(\mathrm{CosSim}_p(u;v) \in [-1,1]\).  
It measures the directional alignment between \(u\) and \(v\) in the dual geometry.  
When \(p=2\), \(\psi_2(v)=v\) and \(q=2\), so the definition reduces to the standard symmetric cosine similarity. 

For convenience, we define another type of generalized cosine similarity, which is simply induced by the inner product: \(\widetilde{\mathrm{CosSim}} _p(u;v)\triangleq \left\langle \frac{u}{\|u\|_p},\,
\frac{v}{\|v\|_q} \right\rangle\).
For any \(p_y \in \partial h(\mathcal{A}y)\) we have \(\|p_y\|_q \le 1\).  The inner product can be expressed as
\[
\langle p_y, \mathcal{A}x \rangle = \|\mathcal{A}x\|_p \|p_y\|_q \widetilde{\mathrm{CosSim}} _p(\mathcal{A}x; p_y).
\]

Substituting this into \eqref{eq:Df_diff} gives the unified decomposition
\begin{equation*}
\begin{split}
D_f(x; y) =& \|\mathcal{A}x\|_p\|p_y\|_q \bigl( 1 - \widetilde{\mathrm{CosSim}} _p(\mathcal{A}x; p_y) \bigr) \\
& + (1-\|p_y\|_q)\|\mathcal{A}x\|_p.
\end{split}
\end{equation*}

This form holds for all \(\mathcal{A}y\), without case distinction:
\begin{itemize}
\item When \(\mathcal{A}y \neq 0\), we have \(p_y = \psi_p(\mathcal{A}y)/\|\psi_p(\mathcal{A}y)\|_q\), hence \(\widetilde{\mathrm{CosSim}} _p(\mathcal{A}x; p_y) = \mathrm{CosSim}_p(\mathcal{A}x; \mathcal{A}y)\), and the expression reduces to \(\|\mathcal{A}x\|_p \bigl(1 - \mathrm{CosSim}_p(\mathcal{A}x; \mathcal{A}y)\bigr)\);
\item When \(\mathcal{A}y = 0\), we have \(\|p_y\|_q \le 1\), and this value acts as a gating weight that blends pure magnitude minimization (\(\|p_y\|_q = 0\)) with directional alignment (\(\|p_y\|_q = 1\)), naturally adapting from strict guidance to weak or no directional guidance.
\end{itemize}

Therefore, \(D_f(x; y)\) provides a convex, magnitude-invariant mechanism for information transfer, with a generalized cosine-similarity interpretation.

\section{DSE-DConv Feature Operator}
\label{sec:operator}
\begin{figure*}[t]
    \centering
    \includegraphics[width=\linewidth]{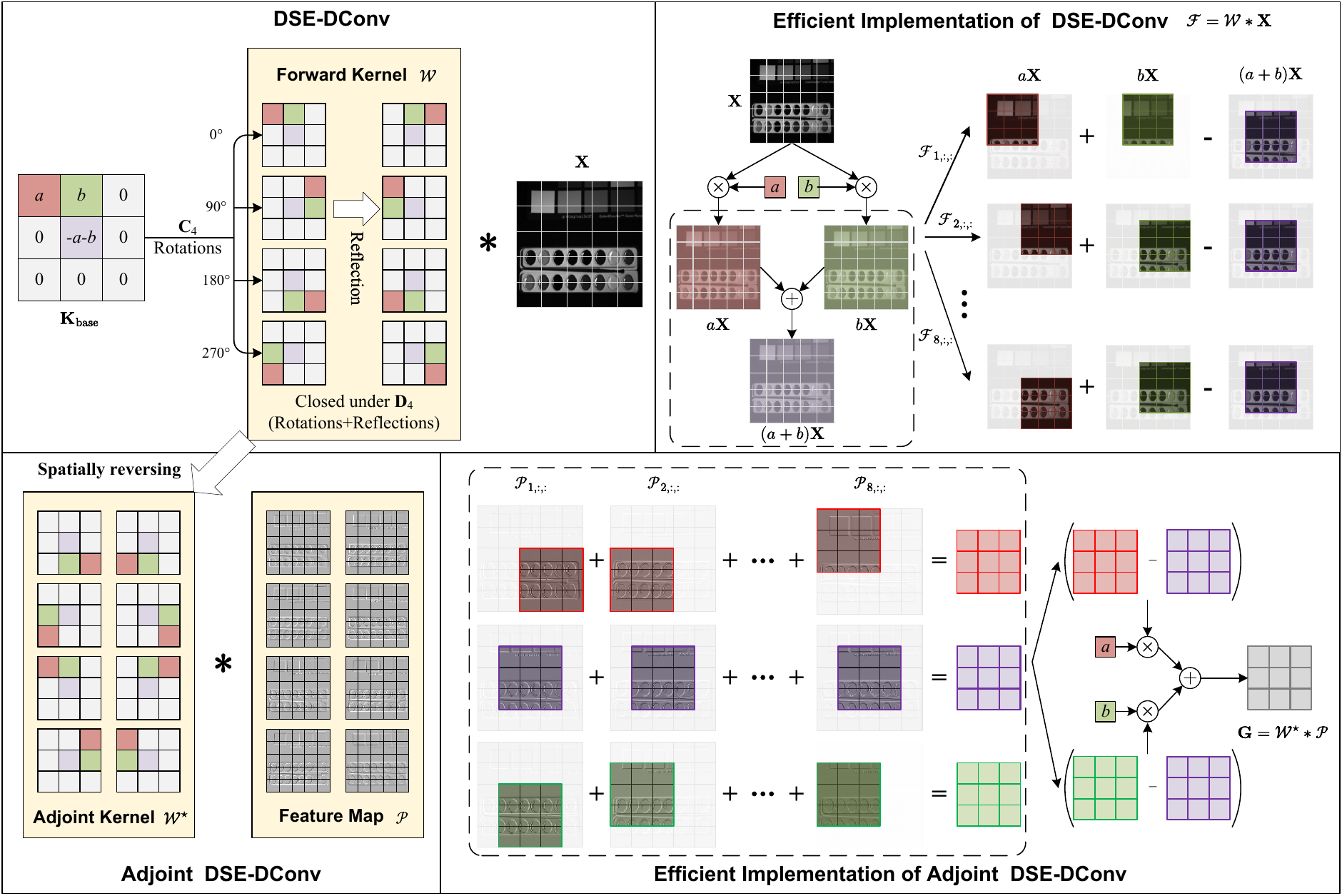}
    \caption{Illustration of the DSE-DConv module and its adjoint operator. The right part shows the speed-optimized implementation with only $2HW$ multiplications.}
    \label{fig:DSE-DConv}
\end{figure*}

The subgradient-guided framework requires a linear operator that extracts structural features whose directions are transferable across spectral bands. We design it as a convolutional layer with kernel \(\mathcal{W}\in\mathbb{R}^{8\times 1\times 3\times 3}\), which naturally captures local spatial structures. Eq.~\eqref{eq:Df_diff} then specializes to
\begin{equation}
R_{\mathcal{W}}(\mathbf{X}; \mathbf{X}_{\text{ref}}) = \sum_{h,w}\|(\mathcal{W} * \mathbf{X})_{:,h,w}\|_p \;-\; \langle \mathcal{P}_{\text{ref}}, \mathcal{W} * \mathbf{X} \rangle,
\label{eq:TSP_msi}
\end{equation}
where \(\mathbf{X}_{\text{ref}}\in\mathbb{R}^{H\times W}\) is a reference image. We set \(p=2\) in our implementation, so the dual subgradient \(\mathcal{P}_{\text{ref}}\) is given by:
\[
(\mathcal{P}_{\text{ref}})_{:,h,w}\in \begin{cases}\left\{\frac{\psi_p((\mathcal{W}*\mathbf{X}_{\text{ref}})_{:,h,w})}{\|\psi_p((\mathcal{W}*\mathbf{X}_{\text{ref}})_{:,h,w})\|}\right\} &, (\mathcal{W}*\mathbf{X}_{\text{ref}})_{:,h,w}\neq\mathbf{0}\\
  \mathcal{B}_2 & ,\text{otherwise}
\end{cases}
\]

\textbf{Remark.} 
When \((\mathcal{W}*\mathbf{X}_{\mathrm{ref}})_{:,h,w}=0\), the pointwise subgradient is not uniquely determined. 
In our algorithm, however, the dual variable is not chosen independently at each pixel, but updated through a convolutionally coupled primal-dual recursion, i.e. Eq.~\eqref{eq:G_dual}. 
It couples neighboring dual variables through the receptive field of \(\mathcal{W}\), so zero-response locations can still receive structural guidance propagated from adjacent nonzero-response regions.

The remainder of this section describes \(\mathcal{W}\).

\subsection{Operator Design}
For MSFA demosaicing, we design the operator with the following properties:

\begin{enumerate}
    \item \textbf{Spectral independence.} 
    The operator is applied to each spectral band independently, so that cross-band information is introduced only through the proposed regularizer rather than through feature extraction itself.

    \item \textbf{\(D_4\)-equivariance.} 
    Since edges and textures may appear in arbitrary orientations, the operator should respond consistently under rotations and reflections. 
    Enforcing \(D_4\)-equivariance reduces orientation bias in the transferred structural guidance.

    \item \textbf{Directional sparsity.} 
    Local structures usually have a dominant orientation. 
    Directionally sparse filters help isolate this orientation and reduce interference between different edge directions.

    \item \textbf{High-pass sensitivity.} 
    MSFA subsampling mainly corrupts high-frequency spatial components. 
    A zero-sum high-pass design therefore directly targets the structures most affected by aliasing.
\end{enumerate}

We meet these requirements by applying all eight \(D_4\) transformations to a two-parameter base kernel that is intrinsically zero-sum and directionally sparse:
\[
\mathbf{K}_{\mathrm{base}} = \begin{bmatrix}
 a &  b & 0 \\
 0 & -a-b & 0 \\
 0 & 0 & 0
\end{bmatrix},
\]
where \(a,b\in\mathbb{R}\) are non-zero parameters. Applying all eight transformations of \(D_4\) to this kernel generates a set of eight directional filters. Together they form \(\mathcal{W}\), which processes each channel independently and thus satisfies all four design requirements. Implementation details are provided in Fig.~\ref{fig:DSE-DConv}.

Owing to parameter sharing and kernel sparsity, the DSE-DConv requires only \(2HW\) multiplications and \(17HW\) additions per channel, substantially fewer than a standard \(3\times3\) convolution. The adjoint operator also enjoys computational savings. Despite memory-access overhead, the runtime speedup over standard convolution remains roughly threefold. 

\subsection{Structural Information Transfer with DSE-DConv}
We next examine why the subgradient with DSE-DConv is suitable for cross-band structure transfer. 
The analysis shows that, for ideal two-region edge patterns, the resulting dual subgradient depends on the edge geometry and polarity, but not on the contrast magnitude. 
This property is consistent with the goal of transferring structure rather than spectral intensity.

Let \(\boldsymbol{v}= (\mathcal{W}*\mathbf{X})_{:,h,w} \in \mathbb{R}^8\) denote the feature vector at position \((h,w)\).
For a binary local patch with intensities \(I_1\) and \(I_2\), we assume without loss of generality that the center pixel is \(I_1\). The DSE-DConv then yields
\[
\boldsymbol{v} = (I_2 - I_1)\,\mathbf{d},
\]
where the \(k\)-th element of \(\mathbf{d}\) satisfies \(d_k \in \{a,\, b,\, a+b\}\). Therefore, \(\mathbf{d}\) depends solely on the binary pattern geometry and the kernel parameters \(a,b\).
Substituting into \eqref{eq:subdiff_p}, the dual subgradient becomes
\[
\mathcal{P}_{:,h,w} = \mathrm{sgn}(I_2-I_1)\,
\frac{\psi_p(\mathbf{d})}{\|\psi_p(\mathbf{d})\|_q}.
\]

Thus, for any binary edge, \(\mathcal{P}\) acts as a constant, scale-free orientation template whose overall sign reflects the polarity of the edge (dark-to-light versus light-to-dark).
This guarantees that the regularizer in Eq.~\eqref{eq:TSP_msi} can transfer stable and interpretable structural information from the reference to the target without being affected by contrast differences.

For patches with three or more distinct intensities, however, the normalized feature vector becomes contrast-dependent. Consequently, the structural information may differ between spectral bands, motivating data-driven pre-training to learn more robust kernels.

\subsection{Self-Supervised Pre-training of DSE-DConv}
Although the constrained kernel design already encodes high-pass, directional, and equivariant properties, the two parameters remain unspecified. 
We therefore pre-train \(\mathcal{W}\) on the 900 training images of the NTIRE 2022 spectral demosaicing dataset. For each spectral cube \(\mathcal{X}\), we randomly permute its band order to obtain a reference cube \(\mathcal{X}_{\text{ref}}\) and minimize the loss based on Eq~.\eqref{eq:TSP_msi}:
\[
\mathcal{L}(\mathcal{W}) = \sum_{c=1}^{C} R_{\mathcal{W}}\bigl(\mathbf{X}_c;\; \mathbf{X}_{\text{ref},c}\bigr).
\]

This loss encourages the learned parameters \(a,b\) to produce feature directions that are as consistent as possible across the spectral dimension.

The parameters are optimized via Adam (learning rate \(10^{-3}\), batch size 16) with early stopping if no improvement is observed for 10 consecutive epochs. Once trained, the operator \(\mathcal{W}\) is frozen and applied to all experiments.

\begin{figure}[ht]
    \centering
    \includegraphics[width=0.95\linewidth]{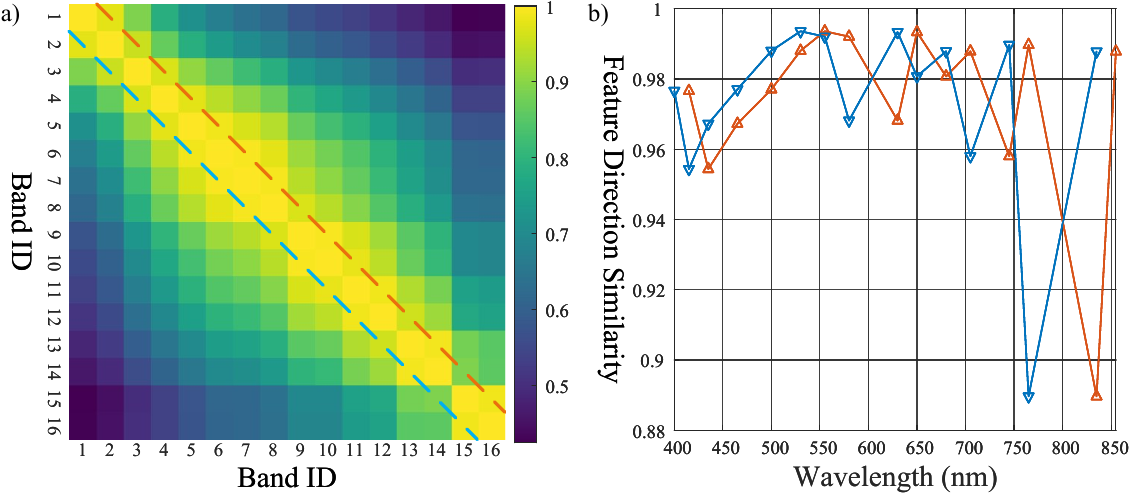}
    \caption{Visualization of feature direction similarity using DSE-DConv. a) Inter-band similarity; b) neighboring-band similarity. The visualization is based on scene 0901 of the NTIRE 2022 spectral demosaicing dataset.}
    \label{fig:cosine}
\end{figure}

Fig.~\ref{fig:cosine} displays the average cosine similarity matrix between all band pairs with learned parameters \(a=-0.5235, b=0.8520\), where each element is defined as
\[
S_{ij} = \frac{1}{N_{ij}} \sum_{h,w} \mathrm{CosSim}_2\bigl( (\mathcal{W} * \mathbf{X}_i)_{:,h,w} ; (\mathcal{W} * \mathbf{X}_j)_{:,h,w} \bigr),
\]
with $N_{ij}$ counting the positions where both feature vectors are non-zero. The similarity matrix exhibits a clear block-diagonal structure, confirming that extracted feature directions are most consistent between spectrally adjacent bands.

\section{Cross-Band Self-Structure Similarity Spectral Demosaicing}
\label{sec:algorithm}
Motivated by the stronger feature-direction similarity between adjacent bands, we instantiate the transfer framework with DSE-DConv, as defined in Eq.~\eqref{eq:TSP_msi}, by regularizing each band using the dual subgradient of its spectral neighbors:
\begin{equation}
R(\mathcal{X}) = \sum_{c=1}^C \bigl( R_{\mathcal{W}}(\mathbf{X}_c;\mathbf{X}_{c-1}) + R_{\mathcal{W}}(\mathbf{X}_c;\mathbf{X}_{c+1}) \bigr).
\label{eq:R_prop}
\end{equation}

To mitigate the influence of residual local directional mismatches and to enforce global spectral coherence, we further augment the objective with a weighted nuclear norm term on the mode-1 unfolding of \(\mathcal{X}\). The full problem is formulated as
\begin{equation}
\min_{\mathcal{X}} \; \frac{1}{2} \|\boldsymbol{y} - \mathbf{\Phi}\boldsymbol{x}\|_F^2 + \lambda \|\mathcal{X}_{(1)}\|_{w,*} + \mu R(\mathcal{X}),
\label{eq:total_obj}
\end{equation}
where \(\lambda\ge 0\) and \(\mu>0\) are regularization parameters.

We now develop an ADMM solver for \eqref{eq:total_obj}. Introducing auxiliary variables \(\mathcal{Z} = \mathcal{X}\), the problem becomes
\begin{align*}
\min_{\mathcal{X}, \mathcal{Z}} \quad & \frac{1}{2} \|\boldsymbol{y} - \mathbf{\Phi}\boldsymbol{x}\|_F^2 + \lambda \|\mathcal{X}_{(1)}\|_{w,*} + \mu R(\mathcal{Z}) \\
\text{s.t.} \quad & \mathcal{X} = \mathcal{Z}.
\end{align*}

The \(\mathcal{Z}\)-subproblem couples each band with its two neighbors in \(R(\mathcal{Z})\). We split it into a forward scan and a backward scan, thereby defining the proximal objectives accordingly:
\begin{align*}
\mathcal{L}_X(\mathcal{X}; \mathcal{Z}, \mathcal{U}) &:= 
\|\boldsymbol{y} - \mathbf{\Phi}\boldsymbol{x}\|_F^2 + \rho\|\mathcal{X} + \mathcal{U} - \mathcal{Z}\|_F^2 + 2\lambda\|\mathcal{X}_{(1)}\|_{w,*}, \\
\mathcal{L}_Z^{\pm}(\mathcal{Z}; \mathcal{X}, \mathcal{U}) &:= 
\frac{1}{2}\|\mathcal{X} + \mathcal{U} - \mathcal{Z}\|_F^2 + \eta \sum_{c=1}^{C} R_{\mathcal{W}}(\mathbf{Z}_c;\mathbf{Z}_{c\mp1}),
\end{align*}
with \(\eta = \mu / \rho\), the ADMM steps form the dual-scan scheme:
\begin{subequations}
\label{eq:dual_scan_steps}
\begin{align}
\left.
\begin{aligned}
\mathcal{X}^{(k+1/2)} 
&= \argmin_{\mathcal{X}} 
\mathcal{L}_X\bigl(\mathcal{X}; \mathcal{Z}^{(k)}, \mathcal{U}^{(k)}\bigr), \\
\mathcal{Z}^{(k+1/2)} 
&= \argmin_{\mathcal{Z}} 
\mathcal{L}_Z^{+}\bigl(\mathcal{Z}; \mathcal{X}^{(k+1/2)}, \mathcal{U}^{(k)}\bigr), \\
\mathcal{U}^{(k+1/2)} 
&= \mathcal{U}^{(k)} + \mathcal{X}^{(k+1/2)} - \mathcal{Z}^{(k+1/2)},
\end{aligned}
\right\} \text{forward} 
\label{eq:forward_scan}
\\
\left.
\begin{aligned}
\mathcal{X}^{(k+1)} 
&= \argmin_{\mathcal{X}} 
\mathcal{L}_X\bigl(\mathcal{X}; \mathcal{Z}^{(k+1/2)}, \mathcal{U}^{(k+1/2)}\bigr), \\
\mathcal{Z}^{(k+1)} 
&= \argmin_{\mathcal{Z}} 
\mathcal{L}_Z^{-}\bigl(\mathcal{Z}; \mathcal{X}^{(k+1)}, \mathcal{U}^{(k+1/2)}\bigr), \\
\mathcal{U}^{(k+1)} 
&= \mathcal{U}^{(k+1/2)} + \mathcal{X}^{(k+1)} - \mathcal{Z}^{(k+1)}.
\end{aligned}
\right\}\text{backward}
\label{eq:backward_scan}
\end{align}
\end{subequations}

\subsection{\texorpdfstring{\(\mathcal{X}\)}--Subproblem: Weighted Low-Rank Regularization}

With auxiliary variables fixed, the \(\mathcal{X}\)-subproblem is
\begin{equation}
\min_{\mathcal{X}} \; \|\boldsymbol{y} - \mathbf{\Phi}\boldsymbol{x}\|_F^2 + \rho \|\mathcal{X} - \mathcal{Z} + \mathcal{U}\|_F^2 + 2\lambda \|\mathcal{X}_{(1)}\|_{w,*}.
\label{eq:X_sub}
\end{equation}

Defining the weight tensor \(\mathcal{D} = \mathcal{M} + \rho\) and the unconstrained minimizer \(\tilde{\mathcal{X}} = (\mathcal{M} \odot \mathcal{Y} + \rho \mathcal{Z} - \rho \mathcal{U}) \oslash \mathcal{D}\), this becomes a weighted least-squares problem with a weighted nuclear norm:
\begin{equation}
\min_{\mathcal{X}} \; \frac{1}{2} \bigl\| \sqrt{\mathcal{D}} \odot (\mathcal{X} - \tilde{\mathcal{X}}) \bigr\|_F^2 + \lambda \|\mathcal{X}_{(1)}\|_{w,*}.
\label{eq:X_weighted}
\end{equation}
We solve it via proximal gradient descent \cite{Beck2009}. With step size \(0<\tau<1/(1+\rho)\), the proximal step applies weighted singular value thresholding (SVT) to the mode-1 unfolding:
\begin{equation}
\mathcal{X}^{[m+1]}_{(1)} = \operatorname{SVT}_{\tau \lambda \boldsymbol{w}} \Bigl( \bigl( \mathcal{X}^{[m]} - \tau \mathcal{D} \odot (\mathcal{X}^{[m]} - \tilde{\mathcal{X}}) \bigr)_{(1)} \Bigr).
\label{eq:prox_X}
\end{equation}

Following~\cite{Candes2010,xie_weighted_2016}, we set \(\lambda=1/\sqrt{\max(H,W)}\) and \(w_i=1/(\sigma_i + 1)^p\), where \(\sigma_i\) are the singular values of \(\tilde{\mathcal{X}}_{(1)}\).

\subsection{\texorpdfstring{\(\mathcal{Z}\)}--Subproblem: Band Sequential Primal-Dual Optimization}

For the forward scan \eqref{eq:forward_scan} (the backward case is analogous), the objective of the \(\mathcal{Z}\)-subproblem is
\begin{equation}
\min_{\mathcal{Z}} \; \frac{1}{2} \|\mathcal{Z} - \mathcal{U} - \mathcal{X}\|_F^2 + \eta \sum_{c=1}^{C} R_{\mathcal{W}}(\mathbf{Z}_c;\mathbf{Z}_{c-1}).
\label{eq:Z_sub}
\end{equation}

We solve it band-by-band using a Gauss-Seidel update. Using the 1-homogeneity as described in Eq.~\eqref{eq:Df_diff}, the subproblem for band \(c\) becomes:
\begin{equation}
\min_{\mathbf{Z}_c} \max_{\mathcal{P}_{c}\in \mathcal{B}} \; \frac{1}{2} \|\mathbf{Z}_c - \mathbf{U}_c - \mathbf{X}_c\|_F^2 + \eta \langle \mathcal{P}_{c}-\mathcal{P}_{c-1}, \mathcal{W} * \mathbf{Z}_c \rangle.
\label{eq:Z_saddle}
\end{equation}
Exchanging \(\min\) and \(\max\) and setting the derivative w.r.t.\ \(\mathbf{Z}_c\) to zero yields
\begin{equation}
\mathbf{Z}_c = \mathbf{U}_c + \mathbf{X}_c + \eta \, \mathcal{W}^{\star} * (\mathcal{P}_{c-1}- \mathcal{P}_{c}),
\label{eq:Z_update}
\end{equation}
which reduces the problem to the dual update for \(\mathcal{P}_c\):
\begin{equation}
\mathcal{P}_{c} = \operatorname*{argmin}_{\mathcal{P}_{c}\in \mathcal{B}} \; \frac{\eta}{2} \|\mathcal{W}^{\star} * (\mathcal{P}_{c} - \mathcal{P}_{c-1})\|_F^2 - \bigl\langle \mathcal{P}_{c}, \; \mathcal{W} * (\mathbf{U}_c + \mathbf{X}_c) \bigr\rangle.
\label{eq:G_dual}
\end{equation}

Inspired by Chambolle's semi-implicit projection scheme~\cite{chambolle_algorithm_2004}, we perform \(K\) inner iterations to obtain an approximate solution of \eqref{eq:G_dual}:
\begin{equation}
(\mathcal{P}_{c}^{[k+1]})_{:,h,w} = \frac{ (\mathcal{P}_{c}^{[k]} - \Delta^{[k]})_{:,h,w}}
{ 1+ \| (\Delta^{[k]})_{:,h,w}\|_2},
\label{eq:G_update}
\end{equation}
where the descent step incorporates a momentum term:
\begin{equation}
\Delta^{[k]} = \tau\nabla \mathcal{L}_{\text{smooth}}^{[k]} - \beta_m \bigl( \mathcal{P}_{c}^{[k]} - \mathcal{P}_{c}^{[k-1]} \bigr),
\label{eq:momentum}
\end{equation}
with \(\nabla \mathcal{L}_{\text{smooth}}^{[k]} = -\mathcal{W} * \bigl( \mathbf{U}_c + \mathbf{X}_c - \eta \, \mathcal{W}^{\star} * (\mathcal{P}_{c}^{[k]} - \mathcal{P}_{c-1}) \bigr)\). Once \(\mathcal{P}_c\) is updated, \(\mathbf{Z}_c\) is recovered via \eqref{eq:Z_update}.

For the boundary condition, we initialize \(\mathcal{P}_0\) from a panchromatic (PAN) estimate \(\mathbf{X}_{0} = \frac{1}{C}\sum_{c=1}^C \mathbf{X}_c\), and update \(\mathcal{P}_0\) at each outer iteration via the dual update procedure as described by Eq.\eqref{eq:G_dual}. For the backward scan we set \(\mathcal{P}_{C+1} = \mathcal{P}_C\), as the forward pass has already propagated reliable structural information through the stack.

\begin{algorithm}[t]
\caption{S3D-Mosaic via ADMM}
\label{alg:S3D}
\begin{algorithmic}[1]
\REQUIRE $\mathbf{Y}$, $\mathcal{M}$, pre-trained $\mathcal{W}$, $\mu,\lambda\ge0$, $\rho_0>0$, $\gamma_\rho>1$, $K_z$, $K_x$, $T$.
\ENSURE Reconstructed MSI $\mathcal{X}$.
\STATE Initialize $\mathcal{Z},\mathcal{U}$ and \(\mathcal{P}_{-1}\) with zeros; set $\eta_0 \gets \mu/\rho_0$.
\STATE Initialize \(\mathcal{X}\gets (\mathcal{M}\odot\mathcal{Y})\oslash (\mathcal{M+\rho})\).
\FOR{$t=0,\dots,T-1$}
    \STATE \textit{// Boundary condition}
    \STATE Compute $\mathbf{P}_0$ from $\mathbf{X}_{0}\!=\!\frac{1}{C}\!\sum_c\mathbf{X}_c$ via \eqref{eq:G_dual}.
    \STATE \textit{// Forward scan}
    \STATE Sequentially obtain $\{\mathbf{Z}_c,\mathbf{P}_c\}_{c=1}^C$ by solving \eqref{eq:Z_saddle} with reference $\mathbf{P}_{c-1}$, using $K_z$ primal-dual iterations \eqref{eq:G_update}-\eqref{eq:momentum}.
    \STATE \(\mathcal{U}\gets\mathcal{U}+\mathcal{X}-\mathcal{Z}\).
    \STATE Update \(\mathcal{X}\) by solving \eqref{eq:X_weighted} via \(K_x\) iterations of SVT.
    \STATE \textit{// Backward scan}
    \STATE Set $\mathbf{P}_{C+1}\!=\!\mathbf{P}_C$; sequentially obtain $\{\mathbf{Z}_c,\mathbf{P}_c\}_{c=C}^1$ with reference $\mathbf{P}_{c+1}$.
    \STATE \(\mathcal{U}\gets\mathcal{U}+\mathcal{X}-\mathcal{Z}\).
    \STATE Update $\mathcal{X}$ as above.
    \STATE \textit{// Adaptive penalty update}
    \IF{$(t+1)\bmod 10 = 0$}
        \STATE $\rho_0 \!\gets\! \rho_0\gamma_\rho$, $\eta_0 \!\gets\! \eta_0/\gamma_\rho$, $\mathcal{U}\!\gets\!\mathbf{0}$.
    \ENDIF
\ENDFOR
\RETURN $\mathcal{X}$.
\end{algorithmic}
\end{algorithm}

The overall procedure is summarized in Algorithm~\ref{alg:S3D}. 
The forward and backward scans alternately propagate magnitude-robust structural directions across the spectral dimension, while the \(\mathcal{X}\)-updates enforce data consistency and optional spectral low-rankness. 
When a high-resolution panchromatic image is available, it can replace the averaged boundary estimate \(\mathbf{X}_{0}\), providing stronger structural initialization and naturally extending the framework to pan-mosaic fusion.

\section{Experimental Results and Discussion}
\label{sec:experiments}
In this section, we conduct comprehensive experiments to evaluate the proposed method. We first detail the experimental setup, then present and discuss results on spectral demosaicing, pan-mosaic fusion, ablation studies, and real-world validation.

\subsection{Experimental Setup}

\textbf{CAVE dataset}: The CAVE dataset consists of indoor laboratory scenes with spectral image dimensions of $512\times512\times31$~\cite{yasuma_generalized_2010}. Following the conventional protocol in the literature~\cite{chen_unsupervised_2025}, we adopt a $5\times5$ MSFA: a $500\times500$ sub-image is cropped from the top-left corner, and the first 25 of the 31 bands are selected.

For the pan-mosaic fusion task, we follow~\cite{wang_equivariant_2026, dian_equivariant_2026}: bands 12 to 27 of the original image are selected to form a 16-band hyperspectral image. The mosaic observation is then obtained by $8\times8$ mosaicking and $2\times2$ average downsampling. The PAN image is generated using an artificially defined spectral response function $R = (1, 1, 2, 4, 8, 9, 10, 12, 16, 12, 10, 9, 7, 3, 2, 1)/107$.

\textbf{NTIRE 2022 spectral demosaicing dataset}: NTIRE 2022 contains 1000 full-resolution MSIs 
of natural scenes with 16 bands at $480\times512$ spatial resolution, officially split into 900 training, 50 validation, and 50 testing. The raw images are generated with a \(4\times4\) MSFA. As the test set lacks ground truth, all methods are evaluated on the official validation set.

\textbf{Real-world dataset}: To evaluate generalization under real imaging conditions, we collected real-captured raw images using a UAV-mounted Specvision camera equipped with a \(3\times 3\) MSFA. The acquisition protocol is detailed in Section~\ref{sec:real}.

\textbf{Evaluation metrics}: We employ four full-reference image quality metrics to evaluate reconstruction performance on synthetic datasets: the PSNR, the SSIM, the SAM, and the ERGAS. ERGAS is used only for the pan-mosaic fusion evaluation. For real-world datasets, we evaluate performance using the NIQE on false-color composites.

All the hyperparameters listed in Table~\ref{tab:hyperparams} were kept identical across every experiment reported in this paper, including CAVE demosaicing, NTIRE 2022 demosaicing, CAVE pan-mosaic fusion, and the real-world Specvision UAV dataset.
\begin{table}[htbp]
  \centering
  \caption{Hyperparameter settings for the proposed methods. All parameters are fixed across datasets.}
  \label{tab:hyperparams}
  \begin{tabular}{lll}
    \toprule
    Parameter & S3D-Mosaic & S3D-Mosaic\(^*\) \\
    \midrule
    ADMM iterations \(T\)                   & 100              & 100                \\
    Initial penalty $\rho_0$           & $1\times 10^{-3}$& $1\times 10^{-3}$  \\
    Initial structure weight $\mu_0$   & $1\times 10^{-3}$& $1\times 10^{-3}$  \\
    Penalty growth factor $\gamma_\rho$      & 1.8              & 1.8                \\
    $\rho$ update interval             & 10               & 10                 \\
    Z-subproblem FPI iterations \(K_z\)          & 45               & 45                 \\
    Low-rank penalty $\lambda$         & 0                & \(1/\sqrt{\max(H,W)}\)    \\
    X-subproblem PGD steps \(K_x\)          & 0                & 7                  \\
    \bottomrule
  \end{tabular}
\end{table}

We compare the proposed method with three traditional algorithms: WB~\cite{brauers_color_2006}, ItSD~\cite{mizutani_multispectral_2014} and PPID~\cite{mihoubi_multispectral_2017} as well as two state-of-the-art unsupervised deep learning methods, USD~\cite{feng_unsupervised_2024} and UnNull~\cite{chen_unsupervised_2025}. For the pan-mosaic fusion task, we additionally include PPID-PanGAN, LSAN-PanGAN, LSAN-WFANet, EFN~\cite{wang_equivariant_2026}, and EBVIF~\cite{dian_equivariant_2026}. 
All competing methods are executed using the authors' original code with their recommended hyperparameters, with two exceptions: Since UnNull does not provide per-scene tuning parameters for the NTIRE 2022 dataset, we adopt the hyperparameters from its implementation for real-data scenarios. USD was originally trained on ICVL with a $5\times5$ MSFA configuration and performs poorly when tested on CAVE; we therefore retrain it on CAVE excluding the test scenes while keeping all other hyperparameters identical to the ICVL setup.

\subsection{Comparison with Spectral Demosaicing Methods}

\paragraph{Results and Discussion on CAVE Dataset}
\begin{table*}[htbp]
  \centering
  \caption{Quantitative comparison on the CAVE spectral demosaicing dataset. Best and second best are \textbf{bold} and \underline{underlined}, respectively}
  \label{tab:comparison_on_cave5x5}
  \resizebox{0.9\linewidth}{!}{\begin{tabular}{ccccccccccccc}
    \toprule
    Metrics & Methods & balloons & beads & cd    & chart & clay  & cloth & bears & feathers & flowers & oil   & AVG \\
    \midrule
    \multirow{7}[2]{*}{PSNR \textuparrow} & WB    & 36.75  & 24.93  & 35.70  & 27.25  & 36.83  & 27.18  & 32.96  & 28.94  & 34.73  & 30.63  & 31.59  \\
          & ItSD  & 37.12  & 23.37  & 32.55  & 30.55  & 34.69  & 31.53  & 36.17  & 31.46  & 35.26  & 32.34  & 32.50  \\
          & PPID  & 38.89  & 26.24  & 36.60  & 29.00  & 38.34  & 28.54  & 35.61  & 31.00  & 37.12  & 31.05  & 33.24  \\
          & USD (TIP, 2024) & 40.93  & 27.22  & 36.64  & 31.29  & 39.87  & 30.93  & 38.74  & 32.64  & 38.76  & 32.47  & 34.95  \\
          & UnNull (TIP, 2025) & 43.66  & \textbf{31.28} & 36.41  & 34.93  & 42.89  & 34.44  & 41.63  & 37.38  & 41.01  & 34.55  & 37.82  \\
          & S3D-Mosaic & \textbf{46.61} & 31.09  & \textbf{37.71} & \underline{37.73} & \underline{45.72} & \underline{36.02} & \underline{43.98} & \underline{38.80} & \underline{45.04} & \underline{36.05} & \underline{39.88} \\
          & S3D-Mosaic* & \underline{46.22} & \underline{31.20} & \underline{37.35} & \textbf{38.19} & \textbf{46.20} & \textbf{36.41} & \textbf{44.11} & \textbf{39.15} & \textbf{45.38} & \textbf{36.22} & \textbf{40.04} \\
    \midrule
    \multirow{7}[2]{*}{SAM \textdownarrow} & WB    & 3.640  & 12.358  & 6.712  & 9.045  & 9.349  & 11.793  & 2.262  & 7.628  & 9.640  & 10.307  & 8.273  \\
          & ItSD  & 4.595  & 18.385  & 5.908  & 10.037  & 10.269  & 10.215  & 2.194  & 9.411  & 11.815  & 10.074  & 9.290  \\
          & PPID  & 3.524  & 12.043  & 6.511  & 8.684  & 9.330  & 10.674  & 2.026  & 7.388  & 9.413  & 10.383  & 7.998  \\
          & USD (TIP, 2024) & 2.995  & 10.384  & 6.095  & 7.525  & 8.994  & 8.368  & 1.570  & 6.085  & 8.915  & 8.887  & 6.982  \\
          & UnNull (TIP, 2025) & 3.010  & 8.449  & 6.887  & 7.254  & 8.925  & 7.080  & 1.440  & 5.584  & 8.631  & 7.897  & 6.516  \\
          & S3D-Mosaic & \textbf{2.137} & \underline{5.837} & \underline{4.567} & \textbf{5.167} & \underline{6.696} & \underline{5.116} & \underline{1.131} & \underline{3.885} & \textbf{5.979} & \underline{5.905} & \underline{4.642} \\
          & S3D-Mosaic* & \underline{2.229} & \textbf{5.707} & \textbf{4.554} & \underline{5.192} & \textbf{6.554} & \textbf{4.806} & \textbf{1.098} & \textbf{3.794} & \underline{6.181} & \textbf{5.768} & \textbf{4.588} \\
    \midrule
    \multirow{7}[2]{*}{SSIM \textuparrow} & WB    & 0.974  & 0.798  & 0.949  & 0.880  & 0.967  & 0.681  & 0.955  & 0.903  & 0.930  & 0.782  & 0.882  \\
          & ItSD  & 0.963  & 0.652  & 0.934  & 0.897  & 0.931  & 0.861  & 0.951  & 0.896  & 0.921  & 0.859  & 0.887  \\
          & PPID  & 0.981  & 0.843  & 0.953  & 0.910  & 0.971  & 0.762  & 0.967  & 0.929  & 0.953  & 0.773  & 0.904  \\
          & USD (TIP, 2024) & 0.986  & 0.875  & 0.955  & 0.937  & 0.976  & 0.850  & 0.974  & 0.947  & 0.961  & 0.824  & 0.928  \\
          & UnNull (TIP, 2025) & 0.989  & 0.933  & 0.946  & 0.961  & 0.980  & 0.926  & 0.980  & 0.970  & 0.973  & 0.906  & 0.956  \\
          & S3D-Mosaic & \textbf{0.993} & \underline{0.948} & \textbf{0.973} & \textbf{0.984} & \textbf{0.991} & \underline{0.953} & \textbf{0.987} & \underline{0.983} & \textbf{0.990} & \underline{0.931} & \underline{0.973} \\
          & S3D-Mosaic* & \underline{0.992} & \textbf{0.949} & \underline{0.972} & \textbf{0.984} & \textbf{0.991} & \textbf{0.956} & \textbf{0.987} & \textbf{0.984} & \textbf{0.990} & \textbf{0.933} & \textbf{0.974} \\
    \bottomrule
  \end{tabular}}
\end{table*}
As shown in Table~\ref{tab:comparison_on_cave5x5}, traditional methods (WB, ItSD, PPID) are clearly limited in both spatial and spectral fidelity, especially on texture-rich scenes such as \textit{beads} and \textit{cloth}. The unsupervised learning-based methods USD and UnNull significantly improve the reconstruction quality, but still exhibit some spectral distortion on certain scenes.
S3D-Mosaic and S3D-Mosaic* substantially outperform all baselines, with average PSNR gains of over 2 dB against UnNull and more than 4 dB against USD. The consistent reduction in SAM indicates that the self-structure similarity prior effectively preserves spectral fidelity, and the further gain of S3D-Mosaic* confirms that the low-rank term captures complementary inter-band correlations once spatial structure is reliably established.

\paragraph{Results and Discussion on NTIRE 2022 Dataset}

\begin{table*}[htbp]
  \centering
  \caption{Quantitative comparison on the NTIRE 2022 spectral demosaicing dataset. Best and second best are \textbf{bold} and \underline{underlined}, respectively.}
  \label{tab:comparison_on_arad4x4}
  \resizebox{0.85\linewidth}{!}{\begin{tabular}{cccccccc}
    \toprule
    Metrics & UnNull (TIP, 2025) & WB    & ItSD  & PPID  & USD (TIP, 2024)   & S3D-Mosaic & S3D-Mosaic* \\
    \midrule
    PSNR \textuparrow & 29.68±4.10 & 35.00±3.54 & 34.96±3.37 & 37.71±3.53 & 43.39±3.58 & \underline{43.97}±3.91 & \textbf{44.75}±4.10 \\
    SAM \textdownarrow & 9.455±0.854 & 3.800±1.404 & 4.413±1.417 & 3.263±1.339 & 2.223±0.960 & \underline{1.876}±0.867 & \textbf{1.722}±0.819 \\
    SSIM \textuparrow & 0.850±0.082 & 0.922±0.060 & 0.931±0.030 & 0.953±0.037 & 0.981±0.014 & \underline{0.985}±0.010 & \textbf{0.987}±0.010 \\
    \bottomrule
  \end{tabular}}
\end{table*}

Table~\ref{tab:comparison_on_arad4x4} shows that USD achieves strong performance with 43.39 dB PSNR and 2.223\textdegree SAM, substantially outperforming traditional approaches. In contrast, UnNull, despite its reported success on CAVE, drops to 29.68 dB on this 4\(\times\)4 MSFA dataset. This drop is likely caused by two factors: (i) its PixelShuffle upsampler may introduce checkerboard artifacts that are amplified by even upscaling factors and alias with spectral replicas~\cite{agnihotri_improving_2025}; (ii) its online DIP paradigm requires per-image tuning~\cite{liang_analysis_2025, alkhouri_understanding_2026}, and the fixed configuration fails to generalize across the 50 validation scenes. 
S3D-Mosaic* achieves the highest PSNR (44.75 dB) and lowest SAM (1.722°) without any adjustment. Notably, the nuclear norm term further reduces the standard deviation of SAM from 0.867° to 0.819°, suggesting that it not only improves the average spectral accuracy but also helps suppress outlier degradations on scenes with weak directional structure.

\begin{table*}[htbp]
  \centering
  \caption{Quantitative comparison on the CAVE pan-mosaic fusion dataset. Best and second best are \textbf{bold} and \underline{underlined}, respectively}
  \label{tab:comparison_on_cave_dc}
    \resizebox{0.95\linewidth}{!}{\setlength{\tabcolsep}{2pt}\begin{tabular}{ccccccccccccccc}
\toprule
    Metrics & Methods & beans & oil   & paints & photo & pompoms & apples & peppers & sponges & toys  & superballs & spools & watercolors & AVG \\
    \midrule
    \multirow{7}[2]{*}{PSNR \textuparrow} & PPID-PanGAN & 32.37  & 34.90  & 32.22  & 38.66  & 33.81  & 39.34  & 38.76  & 31.19  & 33.46  & 35.36  & 35.38  & 32.77  & 34.85  \\
          & LSAN-WFANet & 30.16  & 33.05  & 30.89  & 38.04  & 36.32  & 41.25  & 39.23  & 33.70  & 35.88  & 37.79  & 37.48  & 29.88  & 35.31  \\
          & LSAN-PanGAN & 32.47  & 35.44  & 33.01  & 39.73  & 37.87  & 43.06  & 41.15  & 34.01  & 36.98  & 39.02  & 39.86  & 32.70  & 37.11  \\
          & EFN (TIP, 2026) & 36.98  & 37.67  & 37.56  & 41.66  & 39.25  & 47.09  & 46.30  & 38.87  & 39.52  & 42.48  & 43.26  & 36.05  & 40.56  \\
          & EBVIF (TPAMI, 2026) & 37.56  & 38.59  & 36.47  & 41.83  & 41.88  & 45.79  & 45.47  & 39.25  & 41.23  & 41.53  & 43.39  & 38.14  & 40.93  \\
          & S3D-Mosaic & \textbf{39.84} & \textbf{39.81} & \underline{39.74} & \textbf{42.51} & \textbf{42.84} & \textbf{50.08} & \textbf{49.91} & \underline{42.33} & \textbf{44.16} & \textbf{46.04} & \textbf{45.50} & \textbf{40.39} & \textbf{43.60} \\
          & S3D-Mosaic* & \underline{39.79} & \underline{39.55} & \textbf{39.99} & \underline{42.10} & \underline{42.46} & \underline{48.05} & \underline{48.07} & \textbf{42.63} & \underline{43.77} & \underline{45.71} & \underline{45.14} & \underline{40.08} & \underline{43.11} \\
    \midrule
    \multirow{7}[2]{*}{SAM \textdownarrow} & PPID-PanGAN & 14.754  & 10.702  & 9.250  & 15.323  & 7.921  & 10.623  & 8.513  & 5.864  & 13.593  & 12.483  & 14.357  & 7.393  & 10.898  \\
          & LSAN-WFANet & 11.894  & 8.254  & 7.679  & 9.295  & 5.547  & 6.375  & 5.470  & 3.959  & 8.751  & 9.582  & 8.816  & 6.107  & 7.644  \\
          & LSAN-PanGAN & 12.132  & 8.381  & 7.295  & 9.537  & 5.260  & 6.951  & 5.807  & 4.285  & 8.287  & 10.212  & 9.259  & 5.958  & 7.780  \\
          & EFN (TIP, 2026) & 6.962  & 6.018  & 4.939  & 6.811  & 4.034  & 4.633  & 3.731  & 2.716  & 6.078  & 7.304  & 6.275  & 4.280  & 5.315  \\
          & EBVIF (TPAMI, 2026) & 9.933  & 7.446  & 6.298  & 10.912  & 4.231  & 6.807  & 5.065  & 3.377  & 9.305  & 9.511  & 8.985  & 5.440  & 7.276  \\
          & S3D-Mosaic & \underline{4.354} & \underline{4.571} & \textbf{3.278} & \textbf{4.805} & \underline{2.634} & \textbf{3.554} & \textbf{2.757} & \textbf{1.678} & \textbf{3.376} & \textbf{5.217} & \textbf{4.456} & \textbf{3.025} & \textbf{3.642} \\
          & S3D-Mosaic* & \textbf{4.174} & \textbf{4.501} & \underline{3.436} & \underline{5.004} & \textbf{2.581} & \underline{4.307} & \underline{3.798} & \underline{1.794} & \underline{3.561} & \underline{5.559} & \underline{4.629} & \underline{3.069} & \underline{3.868} \\
    \midrule
    \multirow{7}[2]{*}{SSIM \textuparrow} & PPID-PanGAN & 0.899  & 0.936  & 0.952  & 0.956  & 0.936  & 0.967  & 0.971  & 0.955  & 0.940  & 0.939  & 0.940  & 0.958  & 0.946  \\
          & LSAN-WFANet & 0.917  & 0.915  & 0.945  & 0.972  & 0.940  & 0.986  & 0.981  & 0.932  & 0.961  & 0.962  & 0.969  & 0.917  & 0.950  \\
          & LSAN-PanGAN & 0.934  & 0.954  & 0.966  & 0.974  & 0.967  & 0.986  & 0.986  & 0.971  & 0.972  & 0.967  & 0.977  & 0.966  & 0.968  \\
          & EFN (TIP, 2026) & 0.974  & 0.964  & 0.983  & 0.985  & 0.977  & 0.995  & 0.994  & 0.987  & 0.987  & 0.989  & 0.989  & 0.977  & 0.983  \\
          & EBVIF (TPAMI, 2026) & 0.958  & 0.961  & 0.973  & 0.978  & 0.978  & 0.989  & 0.990  & 0.979  & 0.978  & 0.975  & 0.982  & 0.973  & 0.976  \\
          & S3D-Mosaic & \underline{0.987} & \textbf{0.973} & \textbf{0.991} & \textbf{0.987} & \textbf{0.984} & \textbf{0.997} & \textbf{0.997} & \textbf{0.991} & \textbf{0.993} & \textbf{0.994} & \textbf{0.992} & \textbf{0.988} & \textbf{0.990} \\
          & S3D-Mosaic* & \textbf{0.988} & \underline{0.972} & \textbf{0.991} & \underline{0.986} & \textbf{0.984} & \underline{0.995} & \underline{0.995} & \underline{0.990} & \underline{0.992} & \underline{0.993} & \textbf{0.992} & \underline{0.987} & \underline{0.989} \\
    \midrule
    \multirow{7}[2]{*}{ERGAS \textdownarrow} & PPID-PanGAN & 4.458  & 3.406  & 3.427  & 5.038  & 2.935  & 4.842  & 2.545  & 2.581  & 4.665  & 8.511  & 5.086  & 1.521  & 4.085  \\
          & LSAN-WFANet & 5.329  & 3.932  & 3.943  & 5.393  & 2.028  & 3.918  & 2.341  & 1.791  & 3.420  & 6.255  & 3.917  & 2.067  & 3.695  \\
          & LSAN-PanGAN & 4.160  & 3.158  & 3.156  & 4.558  & 1.764  & 3.286  & 1.956  & 1.999  & 3.100  & 5.613  & 3.160  & 1.513  & 3.119  \\
          & EFN (TIP, 2026) & 2.872  & 2.753  & 2.061  & 3.813  & 1.529  & 2.189  & 1.254  & 1.068  & 2.476  & 3.679  & 2.240  & 1.134  & 2.256  \\
          & EBVIF (TPAMI, 2026) & 2.643  & 2.514  & 2.327  & 3.902  & 1.254  & 2.517  & 1.304  & 1.078  & 2.023  & 4.528  & 2.304  & 0.905  & 2.275  \\
          & S3D-Mosaic & \underline{1.857} & \underline{2.167} & \underline{1.542} & \textbf{3.698} & \textbf{1.059} & \textbf{1.499} & \textbf{0.728} & \underline{0.685} & \textbf{1.386} & \underline{2.596} & \underline{1.811} & \textbf{0.685} & \textbf{1.643} \\
          & S3D-Mosaic* & \textbf{1.831} & \textbf{2.145} & \textbf{1.475} & \underline{3.785} & \underline{1.067} & \underline{1.795} & \underline{0.826} & \textbf{0.650} & \underline{1.405} & \textbf{2.537} & \textbf{1.747} & \underline{0.688} & \underline{1.663} \\
    \bottomrule
    \end{tabular}}
\end{table*}

\subsection{Comparison with Pan-Mosaic Fusion Methods}
Table~\ref{tab:comparison_on_cave_dc} compares the methods on the pan-mosaic fusion task. A clear trend is that sequential demosaicing-then-pan-sharpening pipelines (PPID-PanGAN, LSAN-WFANet, LSAN-PanGAN) consistently leave high spectral errors, especially on scenes with fine textures such as beans and pompoms. This indicates that spectral details lost during the initial demosaicing stage cannot be recovered by subsequent pan-sharpening. The EI-based methods EFN and EBVIF improve the spatial-spectral balance, yet their SAM values still average above 5\textdegree, suggesting that implicit geometric consistency alone is insufficient to recover the spectral information.

Our S3D-Mosaic, relying solely on the self-structure similarity prior with a high-resolution boundary condition, attains the best performance across all four metrics and achieves an average SAM as low as 3.64\textdegree. Notably, in this fusion setting, S3D-Mosaic outperforms its low-rank augmented variant S3D-Mosaic* on average. This reversal reveals that when a high-quality PAN reference supplies an accurate structural boundary condition, the additional nuclear norm regularizer may become redundant or even slightly detrimental: its global spectral correlation can over-regularize the already well-recovered local spectral variations, introducing subtle distortions that the pure structure prior avoids. This phenomenon reinforces the primacy of the self-structure similarity prior as the core reconstruction engine and confirms that the low-rank term should be understood as a compensating mechanism whose value manifests only when structural references are uncertain.

\subsection{Ablation Study}
\paragraph{Role of prior components and initialization}
We assess the contribution of each term under hard-splitting (HS) and weighted bilinear (WB) initializations on the NTIRE 2022 validation set.
The regularization terms are switched on and off in a controlled manner, and we decompose S3D-Mosaic* into three components:
\textbf{LR} denotes the low-rank prior,
\textbf{Feat.} denotes the feature sparsity prior (using the DSE-DConv operator without subgradient guidance), 
and \textbf{Sub.} activates the subgradient term.

\begin{table}[htbp]
  \centering
  \caption{Ablation of prior components under different initializations on the NTIRE 2022 validation set.}
  \begin{tabular}{ccccccc}
    \toprule
    LR & Feat. & Sub. & Initial & PSNR $\uparrow$ & SAM $\downarrow$ & SSIM $\uparrow$ \\
    \midrule
      &       &       & HS & 16.43 & 76.398 & 0.115 \\
      \checkmark &       &       & HS & 16.43 & 76.398 & 0.115 \\
      & \checkmark &       & HS & 34.56 & 2.987  & 0.897 \\
    \checkmark & \checkmark &       & HS & 34.64 & 2.968  & 0.898 \\
      & \checkmark & \checkmark & HS & 43.97 & 1.876  & 0.985 \\
    \checkmark & \checkmark & \checkmark & HS & \textbf{44.75} & \textbf{1.722}  & \textbf{0.987} \\
    \midrule
      &       &       & WB & 35.00 & 3.800  & 0.922 \\
    \checkmark &       &       & WB & 35.35 & 3.748  & 0.927 \\
      & \checkmark &       & WB & 34.56 & 2.987  & 0.897 \\
    \checkmark & \checkmark &       & WB & 34.64 & 2.968  & 0.898 \\
      & \checkmark & \checkmark & WB & 44.03 & 1.873  & 0.985 \\
    \checkmark & \checkmark & \checkmark & WB & \textbf{44.85} & \textbf{1.752} & \textbf{0.987} \\
    \bottomrule
  \end{tabular}%
  \label{tab:ablation_prior}%
\end{table}

Table~\ref{tab:ablation_prior} leads to three observations.
First, without subgradient guidance, the low-rank prior is nearly ineffective (HS: no gain; WB: only +0.35 dB), and the feature prior alone suppresses all high frequencies indiscriminately; adding low-rank on top brings merely 0.08 dB.
Second, activating the subgradient term (Feat. + Sub.) immediately lifts PSNR to ~44 dB and reduces SAM by over 1°, making initialization differences negligible.
Third, on top of this strong spatial prior, the low-rank term contributes a further 0.8 dB and additional SAM reduction, confirming its complementary role in spectral consistency once the structure has been established.

\paragraph{DSE-DConv Design}
To analyze the contribution of each structural property of the proposed DSE-DConv, we conduct an ablation study by systematically enabling or disabling the central difference (CD), directional sparsity (DS), and group equivariance (GE) constraints, where the fully unconstrained case defaults to a standard learnable convolution.

All models are trained on the NTIRE 2022 spectral demosaicing dataset and evaluated on the CAVE spectral demosaicing benchmark. The quantitative results are summarized in Table~\ref{tab:ablation_dsed}, where the last column reports the theoretical multiplication complexity per spatial position in terms of $HW$. 

\begin{table}[htbp]
  \centering
  \caption{Ablation study on the central difference (CD), directional sparsity (DS), and group equivariance (GE) properties. Training is performed on NTIRE 2022 spectral demosaicing and testing on CAVE.}
  \begin{tabular}{ccccccc}
    \toprule
    CD    & DS    & GE    & PSNR $\uparrow$ & SAM $\downarrow$ & SSIM $\uparrow$ & Mult.~Comp. \\
    \midrule  	 	

          &       &       & 34.88 & 7.186 & 0.927 & $72\,HW$ \\
    \checkmark     &       &       & 22.02  & 21.883  & 0.634  & $64\,HW$ \\
          & \checkmark     &       & 32.12  & 10.274  & 0.880  & $24\,HW$ \\
          &       & \checkmark     & 25.48  & 19.971  & 0.645  & $9\,HW$ \\
    \checkmark     & \checkmark     &       & 39.70  & 4.692  & 0.972  & $16\,HW$ \\
    \checkmark     &       & \checkmark     & 21.59  & 23.885  & 0.594  & $8\,HW$ \\
          & \checkmark     & \checkmark     & 28.33  & 15.275  & 0.786  & $3\,HW$ \\
    \checkmark     & \checkmark     & \checkmark     & \textbf{39.88} & \textbf{4.642} & \textbf{0.973} & $\mathbf{2\,HW}$ \\
    \bottomrule
  \end{tabular}
  \label{tab:ablation_dsed}
\end{table}

\begin{figure*}[htbp]
    \centering
    \includegraphics[width=0.93\linewidth]{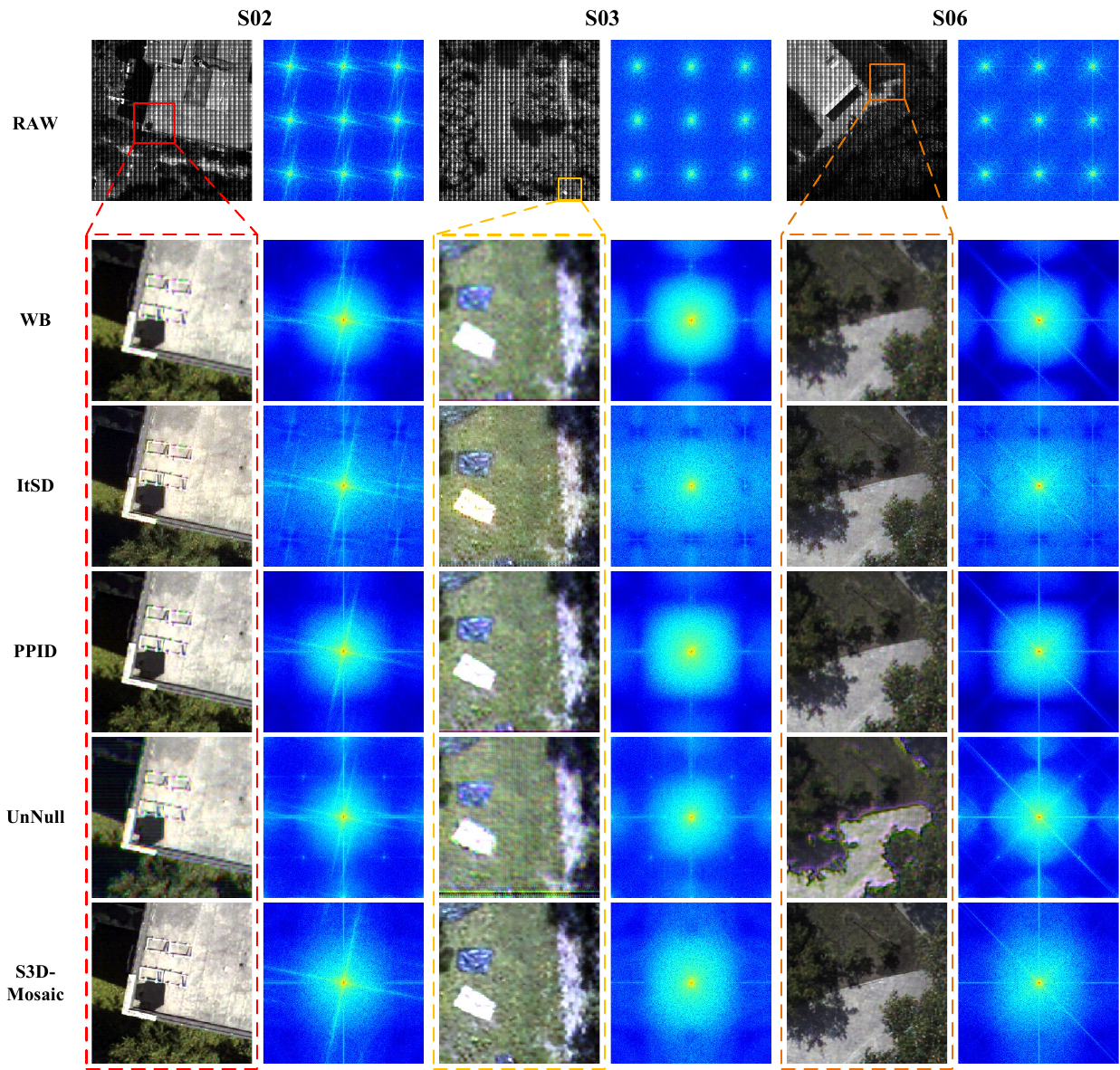}
     \caption{False-color (R:570~nm, G:540~nm, B:454~nm) visual comparison of selected scenes on the real dataset. For each scene, the left column shows a zoomed region indicated in the raw image, and the right column displays the channel-averaged spatial frequency spectrum of the reconstructed scenes.}
     \label{fig:comparison_real}
\end{figure*}
\begin{figure}[htbp]
    \centering
    \includegraphics{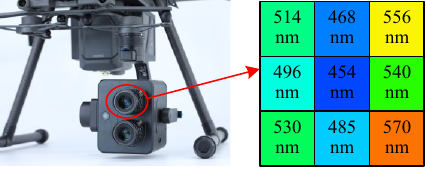}
    \caption{The MSFA layout of the real dataset.}
    \label{fig:real_device}
\end{figure}

Table~\ref{tab:ablation_dsed} reveals a clear synergy.
CD alone suffers from mode collapse without directional constraints; DS alone recovers orientation but retains low-frequency aliasing; GE alone lacks high-pass selectivity.
Pairwise combinations show mixed results: CD + DS achieves 39.70 dB by combining differential priors and diverse orientations, yet it forgoes built-in rotation equivariance; CD + GE collapses because directional sparsity is needed to maintain filter diversity in a compact parameter space; DS + GE benefits from equivariant directionality but still misses high-frequency recovery.
Only when all three constraints are active does the model balance differential extraction, orientation diversity, and transformation robustness, yielding the best performance at the lowest computational cost.

\subsection{Results and Discussion on Real-World Datasets}
\label{sec:real}
We conducted real-world experiments using a Specvision MSFA camera mounted on a UAV platform. The selected camera employs a 3\(\times\)3 spectral filter array, as Fig.~\ref{fig:real_device} shows.

We selected six representative scenes for demosaicing and compared the reconstruction results with WB, ItSD, PPID, and UnNull. Since ground-truth references are unavailable for real-world captures, we evaluated the reconstruction quality using the Naturalness Image Quality Evaluator (NIQE) score on false-color composites, specifically built with spectral channels at R: 570 nm, G: 540 nm, and B: 454 nm.
\begin{table}[htbp]
  \centering
  \caption{Quantitative comparison on the real dataset.}
    \begin{tabular}{ccccccc}
    \toprule
    Methods & S01   & S02   & S03   & S04   & S05   & S06 \\
    \midrule
    WB    & 7.143  & 6.944  & 7.978  & 8.032  & 7.562  & 7.826  \\
    ItSD  & 6.321  & 6.827  & 7.634  & 7.439  & 6.907  & 6.130  \\
    PPID  & 6.141  & 5.733  & 6.440  & 6.144  & 5.944  & 4.725  \\
    UnNull & 6.257  & 6.154  & 6.368  & 5.525  & 4.804  & 4.531  \\
    S3D-Mosaic & \textbf{4.378} & \textbf{4.725} & \textbf{4.941} & \textbf{4.816} & \textbf{4.289} & \textbf{3.834} \\
    \bottomrule
    \end{tabular}%
  \label{tab:comparison_on_real}%
\end{table}

As reported in Table~\ref{tab:comparison_on_real}, the proposed S3D-Mosaic consistently achieves the lowest NIQE scores across all six scenarios, indicating superior perceptual visual quality.

For qualitative assessment, Fig.~\ref{fig:comparison_real} displays zoomed-in false-color patches alongside their corresponding channel-averaged spatial frequency spectra. The false-color visualizations facilitate the identification of edge artifacts and color fringes that are often obscured in single-band intensity images, while the frequency spectra provide insight into the suppression of mosaic-induced aliasing.

A comparative analysis of the spatial frequency responses reveals distinct behaviors among the competing methods. Specifically, WB and PPID function as strong low-pass filters, aggressively suppressing the aliasing peaks caused by mosaic sampling at the expense of attenuating high-frequency components. ItSD preserves more high-frequency details, yet residual aliasing artifacts remain evident. 
UnNull exhibits periodic checkerboard artifacts in its design, which manifest as extra peaks at spatial frequencies corresponding to 2-pixel and 4-pixel periodicities. In contrast, the proposed S3D-Mosaic effectively mitigates the aliasing peaks induced by mosaic sampling while preserving high-frequency details.

\section{Conclusion}
\label{sec:conclusion}
This paper proposed S3D-Mosaic, a model-based method that realizes subgradient-guided cross-band structure transfer for spectral demosaicing. It formulates structural guidance as subgradient-guided feature-direction transfer, designs the compact DSE-DConv operator to extract directional high-pass structures, and incorporates the resulting self-structure similarity prior into a dual-scan ADMM optimization framework. Experimental results on synthetic and real data show that S3D-Mosaic effectively suppresses aliasing, preserves spatial details, and generalizes across different MSFA layouts with fixed hyperparameters.

Future work may proceed in two directions. For spectral demosaicing itself, incorporating contrast-reversal or material-aware mechanisms could improve robustness, adaptive and multi-scale variants of DSE-DConv may enhance representation power, and the optimization can be accelerated via deep unfolding or advanced solvers. More broadly, the subgradient-guided framework provides a magnitude-invariant mechanism for information transfer, naturally extending to image fusion, conditional generation, and inverse problems that exploit self-similarity in feature space; the compact DSE-DConv, which integrates directional sparsity, transformation equivariance, and high-pass selectivity, offers an interpretable and robust low-level feature extractor for lightweight vision systems.

\bibliographystyle{IEEEtran}
\bibliography{IEEEabrv, ref}

\end{document}